\documentclass{article}

\usepackage[final]{neurips_2024}
\usepackage{amssymb}
\usepackage{amsthm}
\usepackage{graphicx}
\usepackage{amsmath} 

\usepackage[utf8]{inputenc} 
\usepackage[T1]{fontenc}    
\usepackage{hyperref}       
\usepackage{url}            
\usepackage{booktabs}       
\usepackage{amsfonts}       
\usepackage{nicefrac}       
\usepackage{microtype}      
\usepackage{xcolor}         

\usepackage{natbib}
\newtheorem{theorem}{Theorem}
\newtheorem{assumption}{Assumption}

\title{General Causal Imputation via Synthetic Interventions}

\author{
  Marco Jiralerspong \And 
  Thomas Jiralerspong \And 
  Vedant Shah \AND
  Dhanya Sridhar\thanks{Canada Cifar AI Chair} \And
  Gauthier Gidel$^*$ \\ \And
  {\normalfont University of Montreal} \\
  Mila - Quebec AI Institute \\
  \texttt{\{marco.jiralerspong,thomas.jiralerspong,vedant.shah,} \\
  \texttt{dhanya.sridhar,gidelgau\}@mila.quebec}
}

\begin{document}

\maketitle

\begin{abstract}
  Given two sets of elements (such as cell types and drug compounds), researchers typically only have access to a limited subset of their interactions. The task of causal imputation involves using this subset to predict unobserved interactions. \citet{squires2022causal} have proposed two estimators for this task based on the synthetic interventions (SI) estimator: SI-A (for actions) and SI-C (for contexts). We extend their work and introduce a novel causal imputation estimator, generalized synthetic interventions (GSI). We prove the identifiability of this estimator for data generated from a more complex latent factor model. On synthetic and real data we show empirically that it recovers or outperforms their estimators. 
  

  
  
\end{abstract}

\section{Introduction}
The problem of determining the result of untested interactions from tested interactions is pervasive in science. Evaluating every possible interaction is often either prohibitively expensive (e.g. examining the effect of thousands of compounds on dozens of different cell types) or unethical (e.g. testing medications on new patient groups). While there are many approaches to this matrix completion problem (increasingly important in unsupervised machine learning), methods that exploit some assumption on the causal structure of these interactions are deemed to be performing \textit{causal imputation}~\cite{squires2022causal}.

Recently, \cite{squires2022causal} have proposed a set of estimators for this particular task. At a high level, the idea of their estimators is a generalization of the idea of synthetic controls \cite{synthetic_controls}. Consider the example of predicting the result of an untested cell type/drug compound pair. Using existing interaction data for this cell type (which they call a \textit{context}), they write the effect of drugs on this cell type as a linear combination of these drugs on other cell types. Applying this linear combination on the drug of interest yields an estimate of the untested combination. This estimator can also be applied in the other direction (i.e. on the drug compounds, which they call \textit{actions}). By assuming the data follows a particular low-rank structure given by a linear factor model, they show their estimators are identifiable.

One potential issue with their estimator is the way they treat the various output dimensions of the interaction. For example, they study the widely-used CMAP (Connectivity Map) dataset, which contain an incomplete collection of interactions between 76 cell types and over 20,000 chemical compounds \cite{subramanian2017next}. For each available interaction, 978 gene expression levels are measured using the L1000 assay (i.e. there is a high-dimensional output for each interaction). Problematically, their estimators use the same linear combination for each dimension of the output: the prediction for the expression of every gene is made using the same linear combination of cell-type expression data. A different linear combination for each dimension of the output could potentially be more appropriate.

\paragraph{Our contributions: } Building on the work of~\citet{squires2022causal}, we propose an extended latent factor model that accounts for the inherent differences between output dimensions. Using this more complex model, we modify their estimator and propose GSI as a novel, more general estimator. We prove the identifiability of this estimator under slightly modified assumptions. Finally, we compare the performance of these estimators on synthetic data and CMAP, showing the efficacy of our method.

\section{Problem Formulation}

\begin{figure}
  \centering
  \includegraphics[width=.70\linewidth]{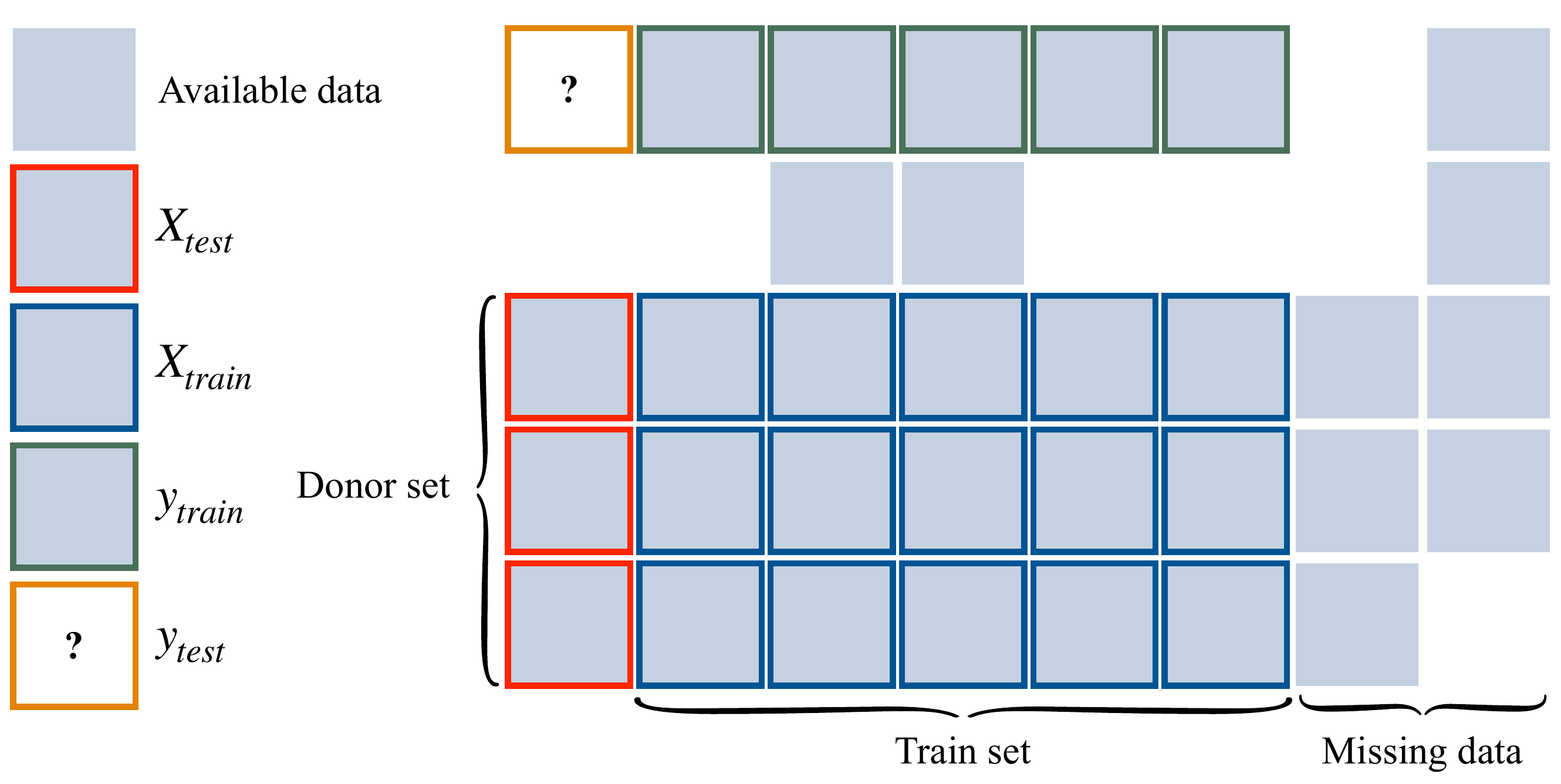}
  \caption{Two-dimensional view of the incomplete matrix and the subsets used by SI-A/SI-C/GSI. The two last columns are missing either donor elements or the target and are thus not included in the train set.}
  \label{fig:sets}

\end{figure}

We begin by defining the causal inference problem of interest. Let $A = \{a_1, \ldots, a_n\}, B = \{b_1, \ldots, b_m \}$ denote the two sets whose interactions we are interested in. We denote by $\mathbf{M} \in \mathbb{R}^{n \times m \times d}$ the \textit{incomplete} tensor of interaction outputs. Its entries $\mathbf{m}_{ij}$ correspond to either the $d$-dimensional result of the interaction of $a_i$ and $b_j$ or $\star$ if this interaction is not available. 

Additionally, we denote by $A(j)$ the set $\{a_i: \mathbf{m}_{ij} \neq \star\}$ and $B(i)$ the set $\{b_i: \mathbf{m}_{ij} \neq \star\}$. Concretely, $A(i)$ is the set of all elements of $A$ for which the interaction with $b_j$ is available (this set is referred to as the donor set by \cite{squires2022causal}) and similarly for $B(i)$. Our goal is to \textbf{infer a missing entry $m_{ij}$ of $M$} by leveraging the non-missing entries of the tensor. 


\subsection{Latent factor models}
In \cite{squires2022causal}, the data in $\textbf{M}$ is assumed to come from the following latent factor model:
\begin{equation}
\label{eq:single_latent}
\mathbf{m_{ij}} = \mathbf{U_i} \mathbf{v_j}
\end{equation}
where $\mathbf{U_i} \in \mathbb{R}^{d \times r}, \mathbf{v_j} \in \mathbb{R}^r$. This model allows each context to have different latent factors for each output dimension. However, for each action, there is \textit{only a single set of latent factors}. There are many practical cases where this assumption is unlikely to hold. For example, consider the effect of various drugs on human proteins. It is likely that the latent factors of the drug affecting binding affinity will be different from those affecting activation strength, reaction speed, etc. To have a single set of latent factors for all these dimensions, the set of latent factors would have to be large which violates the assumptions \cite{squires2022causal} need to show identifiability.

\label{sec:multidim}
To remedy this issue, we propose allowing greater flexibility for the latent factors of the actions with the following linear factor model: 
\begin{equation}
\label{eq:multi_latent}
\mathbf{m_{ij}} = \langle \mathbf{U_{i}}, \mathbf{V_{j}} \rangle \text{ where } \mathbf{U_{i}} \textbf{ and } \mathbf{V_{j}} \in \mathbb{R}^{d \times r}  
\end{equation}
i.e. each dimension of $\mathbf{m_{ij}}$ is equal to the inner product between the corresponding latent factors of the context and the action \textbf{for that dimension}. Interestingly, as a result of this change, the context/actions are now \textbf{symmetric} in the latent factor model: there is nothing differentiating a context from an action. 

This more general formulation says that we have 2 sets of objects $A, B$ and some operation that makes them interact.\footnote{The changes we make to the notation of \cite{squires2022causal} is to illustrate this increase in generality.}  Instead, in \cite{squires2022causal} it is necessary to assign $A$ and $B$ to either contexts or actions (perhaps based on what seems more semantically appropriate). This assignment is sometimes arbitrary and creates undesired asymmetries (for example, when $A = B$).

\section{Related Work}
We review the literature related to synthetic controls (SC), synthetic interventions (SI) and synthetic interventions - action (SI-A). These estimators progressively extend the idea of estimating potential outcomes to more general cases.

\subsection{Synthetic Controls (SC) (\cite{synthetic_controls})}
SC aims to estimate the impact of an intervention on a specific \textit{target} unit. It does this by constructing a \textit{synthetic} control unit from a combination of units that were not intervened upon. The non-intervened units are combined using weights to create a synthetic control unit which is as similar as possible to the target unit. The impact of the intervention can then be estimated by comparing the actual outcome in the target unit with the outcome predicted by the synthetic control.
\subsection{Synthetic Interventions (SI) (\cite{agarwal2023synthetic}}
One limitation of SC is that it can only estimate the impact of the intervention on the intervened-upon units, and not on non-intervened-upon units. SI fixes this by allowing us to predict the impact of any intervention on any unit. SI assumes a pre-intervention period where all units are observed without interventions, followed by a post-intervention period where each unit is intervened upon by one intervention. Given this structure, for a specific intervention $d$ and unit $n$, SI learns a linear model from the pre-intervention data of the units which were intervened upon by $d$ to the pre-intervention data of $n$. The effect of $d$ on $n$ can then be predicted using this same model applied to the post-intervention data of the units which were intervened upon by $d$.

\subsection{Synthetic Interventions - Actions (SI-A) (\cite{squires2022causal}}
One major limitation with SI is its assumption about the structure of the problem. SI-A extends SI to more general observation patterns by allowing potentially \textit{multiple} interventions. It does this by building the linear model along the intervention dimension instead of along the unit dimension, which allows the possibility for multiple interventions to happen to one unit.

\subsection{Matrix Completion (\cite{agarwal2021causal}} 
Matrix completion is the problem of reconstructing a ground-truth matrix from a sparse set of noisy observations. The entries of the matrix are usually assumed to be "missing completely at random" (MCAR). \cite{agarwal2021causal} introduce the Synthetic Nearest Neighbours (SNN) estimator to perform matrix completion and prove the finite-sample consistency and asymptotic normality of their estimator.




\section{Method}
\begin{figure}
  \centering
  \includegraphics[width=\columnwidth]{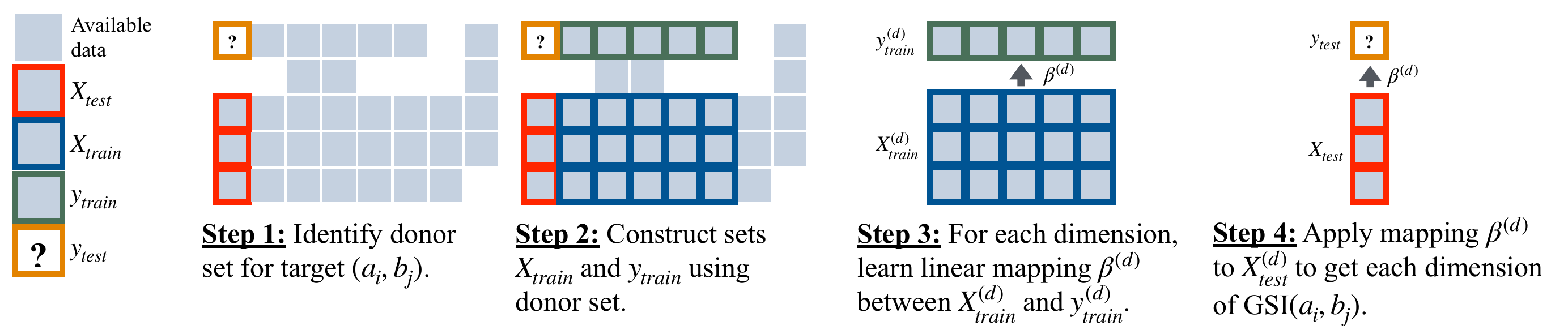}
  \caption{Visualization of the full process for estimating $\textsc{GSI}(a_i, b_j)$. Steps 3 and 4 are performed for each dimension $d$ of the output independently.}
  \label{fig:method}

\end{figure}

We now describe our estimator GSI as a generalization of the estimators SI-A/SI-C proposed by~\citet{squires2022causal}. To simplify notation, for any tensor $\mathbf{x}$, we denote by $\mathbf{x}^{(d)}$ the tensor that results from indexing the tensor's last index at the d-th position (e.g. $\mathbf{M}^{(d)}$ is a $n \times m$ matrix corresponding to the d-th dimension of $\mathbf{M}$).

For a target, $\mathbf{m_{ij}}$, we first identify the donor set $A(j)$ of all elements of $A$ for which the interaction with $b_j$ is available. Then, letting $\mathcal{A} = A(j) \cup \{a_i\}$, we define $\mathcal{B} := \bigcap_{a_i \in \mathcal{A}} B(i)$ as the set of all elements of $B$ for which we have observed interactions between both the donor set and the target $a_i$. Using the data from $\mathcal{B}$, we will learn a mapping from donors to target. This mapping will then be used to generalize from the donor interactions and infer the target value. To do so, we define the following 4 sets:
\begin{align}
\mathbf{X}_{\text{train}} := \mathbf{M}_{A(b),\mathcal{B}}\,, \quad 
\mathbf{y}_{\text{train}} := \mathbf{M}_{a,\mathcal{B}}\,, \quad 
\mathbf{X}_{\text{test}} := \mathbf{M}_{A(b),b}\,, \quad 
\mathbf{y}_{\text{test}} := \mathbf{M}_{a,b}
\end{align}
where indexing by a set is a slight abuse of notation that corresponds to selecting all the indices that match those of the set (see Figure \ref{fig:method}. for an illustration). Here, $\mathbf{X}_{\text{train}}$ and $\mathbf{y}_{\text{train}}$ are order three tensors. To learn a linear mapping from $\mathbf{X}_{\text{train}}$ to $\mathbf{y}_{\text{train}}$, SI-A/SI-C learn the mapping on the flattened tensors. As a result, they only learn a single set of weights which they apply to all dimensions. 

Instead, in GSI, we propose learning a set of weights \textit{per dimension} by simply treating each dimension independently (i.e. each slice of the $\mathbf{X}_{\text{train}}$ and $\mathbf{y}_{\text{train}}$ becomes a regular linear regression problem). Once this mapping is learned, we apply it to each dimension of $\mathbf{X}_{\text{test}}$ and get our estimate $\textsc{GSI}(a_i, b_j)$ of $\mathbf{y}_{\text{test}}$. Concretely, for each dimension $d$:
\begin{align}
    \mathbf{\beta}^{(d)} &= \mathbf{X}_{\text{train}}^{(d)\dagger} \mathbf{y}_{\text{train}}^{(d)}  \label{eq:lin_reg}\\
     \textsc{GSI}(a_i, b_j)^{(d)}&= \hat{\mathbf{y}}_{\text{test}}^{(d)} := \mathbf{X}_{\text{test}}^{(d)} \mathbf{\beta}^{(d)}
\end{align}
where $\mathbf{X}_{\text{train}}^{(d)\dagger}$ denotes the pseudoinverse given by $(\mathbf{X}^T\mathbf{X})^{-1}\mathbf{X}^T$. The entire procedure is illustrated in Figure \ref{fig:method}. Our estimator can also be constructed in the opposite direction \footnote{This change is analogous to the difference between SI-A and SI-C.}, obtaining the donor elements from $B(i)$ and finding the elements of $A$ for which we have data for both the $B(i)$ and $b_j$. We denote this alternative estimator as $\textsc{GSI}(b_j, a_i)$.

In practice, a latent factor model in between (\ref{eq:single_latent}) and (\ref{eq:multi_latent}) is likely to hold (e.g. latent factors are shared between some actions and not others). In such cases, particularly if the data is noisy, imposing some sparsity constraints on the learned $\beta^{(d)}$ is helpful. To do so, instead of (\ref{eq:lin_reg}), we can instead solve the following optimization problem:
\begin{equation}
   \beta = \arg\min_{\beta} \frac{1}{|\mathcal{B}|}\sum_{i=1}^d ||\mathbf{X}_{\text{train}}^{(d)} \beta^{(d)} -  \mathbf{y}_{\text{train}}^{(d)} ||^2_2 + \lambda \sum_{i < j < d} ||\beta^{(i)}  - \beta^{(j)}||_2 \label{eq:regularized_lin_reg}
\end{equation}
where $\lambda$ is a hyperparameter and the regularization term is akin to group lasso regularization, encouraging groups of parameters to be shared between dimensions. We denote by $\textsc{GSI}_\lambda(a_i, b_j)$ the estimator obtained when replacing (\ref{eq:lin_reg}) with the above (with $\textsc{GSI}_0(a_i, b_j)$ corresponding to the original formulation). To solve (\ref{eq:regularized_lin_reg}), we use gradient descent with a decreasing step size, as per \cite{nesterov2004introductory}.

Finally, as mentioned in \cite{squires2022causal}, there is a tradeoff between the size of the donor set and the training examples (the more donor items in $A$ you require, the less likely it is that an element in $B$ will have observed interactions with all the donor items). To address this tradeoff, we add an extra hyperparameter $k$. Instead of selecting $A(j)$ as donor set, we greedily add actions (based on $|B(i)|$) until there are less than $k$ elements in $\mathcal{B}$.

\section{Theory}
We first detail the assumptions we make to show identifiability. These consist of slight modifications to those of \cite{squires2022causal} but we believe they are similarly reasonable.

\begin{assumption}
\label{assu:multi_latent}
The entries of the tensor $\mathbf{M}$ satisfy the multi-dimensional latent factor model defined above where $\mathbf{m_{ij}} = \langle \mathbf{U_i}, \mathbf{V_j} \rangle \text{ with } \mathbf{U_i} \textbf{ and } \mathbf{V_j} \in \mathbb{R}^{d \times r}$.
\end{assumption}
\begin{assumption}
    \label{assu:lin_comb}
    For any target $(i, j)$, we have that the latent factors $\mathbf{U}_i, \mathbf{V}_j$ can be written as a linear combination of the latent factors of the corresponding donor sets:

    \begin{align}
        \exists \lambda_a: \mathbf{U}_i &= \sum_{a \in A(i)} \lambda_a \mathbf{U}_a\qquad
        \label{assu:lin_a}\\
         \label{assu:lin_b}
        \exists \gamma_b: \mathbf{V}_j &= \sum_{b \in B(j)} \gamma_b \mathbf{V}_b
    \end{align}
\end{assumption}

The left equation in~\eqref{assu:lin_b} is equivalent to Assumption 1 of \cite{squires2022causal}. On the other hand, the right equation in~\eqref{assu:lin_b} is similar to Assumption 2 in \cite{squires2022causal}. Notably, it links the elements of $B$ through their latent factors instead of assuming that the link exists at the level of the entries of $\mathbf{M}$ (and thus its rowspan). In practice, it is likely that only Assumption \ref{assu:multi_latent}. is required since, if $r$ is sufficiently small relative to the size of the donor sets, then it is highly likely that Assumption \ref{assu:lin_comb} is satisfied as well.

\begin{theorem}
Suppose $\mathbf{M} \in \mathbb{R}^{N \times M \times D}$ satisfies the high-dimensional latent factor model (Assumption \ref{assu:multi_latent}) with latent factors that satisfy Assumption \ref{assu:lin_comb}. Then, if $\hat{\mathbf{y}}_{\text{test}}^{(d)}$ is derived as above, we have that $ \forall d: \hat{\mathbf{y}}_{\text{test}}^{(d)} = \mathbf{m}_{ij}^{(d)}.$
\end{theorem}
\label{a1:proof}
\begin{proof}
Our proof is inspired by the one of \cite{squires2022causal} but adapted to GSI and with more explicit steps.

Let $ d \in \{1, \ldots, D\}$. As in \cite{squires2022causal}, we first show that the entries of $\mathbf{M}$ can be linked through (\ref{assu:lin_a}). Specifically, we have that $\forall j$:

\begin{align*}
    \mathbf{m}_{aj}^{(d)} &= \langle \mathbf{U_a^{(d)}}, \mathbf{V_j^{(d)}} \rangle \quad \text{ using (\ref{eq:multi_latent})}\\
    &= \left\langle \sum_{i \in A(b)}\lambda_i\mathbf{U_i^{(d)}}, \mathbf{V_j^{(d)}} \right\rangle  \text{ using (\ref{assu:lin_a})}\\
    &= \sum_{i \in A(b)} \lambda_i \left\langle \mathbf{U_i^{(d)}}, \mathbf{V_j^{(d)}} \right\rangle\\
    &= \sum_{i \in A(b)} \lambda_i \mathbf{m}_{ij}^{(d)}.
\end{align*}
A symmetric argument shows that, $\forall i$:
\[
\mathbf{m}_{ib}^{(d)} = \sum_{j \in B(a)} \gamma_j \mathbf{m}_{ij}^{(d)}.
\]
Thus, there exists some set of coefficients $\beta$ (namely the $\lambda_i$) such that $\mathbf{X}_{\text{train}}^{(d)} \beta = \mathbf{y}_{\text{train}}^{(d)}$. The linear regression
in (\ref{eq:lin_reg}) will recover a $\beta$ that satisfies the above.

Note that if the linear regression recovered $\beta_i = \lambda_i, \forall i$, we would be done. However, we cannot guarantee that this solution is unique, only that 
\begin{equation}
    \label{eq:lin_rel}
    \mathbf{m}_{aj}^{(d)} = \sum_{i \in A(b)} \beta_i \mathbf{m}_{ij}^{(d)}
\end{equation}
when regressing over the elements of $A$ or that 
\begin{equation}
 \mathbf{m}_{ib}^{(d)} = \sum_{j \in B(a)} \beta_j \mathbf{m}_{ij}^{(d)}   
\end{equation}
when regressing over the elements of $B$. Using the above, we can now rewrite our estimator as:

\begin{align*}
    \hat{\mathbf{y}}_{\text{test}}^{(d)} &= \mathbf{X}_{\text{test}}^{(d)} \beta^{(d)}\\
    &= \sum_{i \in A(b)} \mathbf{m}_{ib}^{(d)} \beta_i^{(d)} \quad\text{ expanding the matrix multiplication}\\
    &= \sum_{i \in A(b)} \left\langle \mathbf{U_i^{(d)}}, \sum_{j \in B(a)} \gamma_j\mathbf{V_j^{(d)}} \right\rangle \beta_i^{(d)} \quad \text{using (\ref{eq:multi_latent}), (\ref{assu:lin_b}})\\
    &=  \sum_{j \in B(a)} \gamma_j \sum_{i \in A(b)}  \beta_i^{(d)} \left\langle \mathbf{U_i^{(d)}},  \mathbf{V_j^{(d)}} \right\rangle\\
    &= \sum_{j \in B(a)} \gamma_j \sum_{i \in A(b)} \beta_i^{(d)} \mathbf{m}_{ij}^{(d)} \quad \text{using (\ref{eq:multi_latent})}\\
    &= \sum_{j \in B(a)}  \gamma_j \mathbf{m}_{aj}^{(d)} \quad \text{using (\ref{eq:lin_rel})} \\
    &= \mathbf{m}_{ab}^{(d)} \quad \text{using (\ref{assu:lin_b})}
\end{align*}
\end{proof}
\vspace{-2pt}
A symmetric proof exists for regressing over $B$. Essentially, for these estimators to work, two forms of generalization are needed. One first needs to be able to write the target $a_i$ as a linear combination of the $A(b)$. This ensures that it is possible to generalize from the known elements of $A$ to the target. Then, we need to be able to write the target $b_j$ as a linear combination of the $B(a)$. This ensures that the linear combination we learn on $\mathbf{X}_{\text{train}}$ will generalize to $\mathbf{X}_{\text{test}}$.


\section{Experiments \& Results}
\label{sec:exp}

We compare the proposed set of estimators against the baseline estimators used in \cite{squires2022causal}, namely: Mean over Actions, Mean over Contexts, Two Way Mean, Fixed Action Effect in addition to SI-A and SI-C.

\subsection{Synthetic data}
We first evaluate and compare all approaches on multi-dimensional synthetic data generated by the data generating processes described by the latent factor models (\ref{eq:single_latent}) and (\ref{eq:multi_latent}). respectively. For each, we set $|A|$ to 50, $|B|$ to 100 and randomly sample values for their latent factors. We use $d=3$ as the number of dimension of the data (showing the benefits of GSI even for low dimensional data). Table~\ref{res:synth} shows the Mean Absolute Errors (MAE) of all the estimators w.r.t to the ground truth, including the newly proposed GSI.

\begin{table}[h]
\caption{Comparisons of MAE different estimators across singular dimensional and multi-dimensional latent factor models}
\vspace{-0.3cm}
\begin{center}
\resizebox{\textwidth}{!}{
\begin{tabular}{c|cc|cc}
\toprule
                                     & \multicolumn{2}{c}{Original Latent Factor Model}   & \multicolumn{2}{c}{Multidimensional Latent Factor Model}   \\
\midrule
Estimator & Mean &  Standard Deviation & Mean & Standard Deviation  \\
\midrule
Mean over Actions & 1.74 & 1.36 & 3.35 & 2.60  \\
Mean over Contexts & 1.79 & 1.46 & 3.36 & 2.60     \\
Two Way Mean & 1.74 & 1.38 & 3.28 & 2.55  \\
Fixed Action Effect & 2.85 & 2.37 & 4.32 & 3.23  \\
SI-A & 0.81  & 0.87 & 2.32 & 5.60  \\
SI-C & \textbf{0.08} & \textbf{0.32} & 2.82 & 2.42  \\
GSI(a,b) & 0.81 & 0.87 & 1.86 & 1.93  \\
GSI(b,a) & \textbf{0.08} & \textbf{0.32} & \textbf{0.09} & \textbf{0.42}  \\

\bottomrule
\end{tabular}
}
\label{res:synth}
\end{center}
\vspace{-0.3cm}
\end{table}

As expected, on the original latent factor model, the GSI variants perform exactly like SI-A and SI-C respectively, since both the two newly proposed estimators are equivalent to SI-A and SI-C for the original latent factor model . However, in multidimensional cases, the two variants of GSI \textbf{significantly outperform SI-A and SI-C respectively}, with $\textsc{GSI}(b, a)$ performing the best of all with a mean MAE of 0.08. 


\subsection{CMAP}
\begin{figure}
  \centering
  \includegraphics[width=\columnwidth]{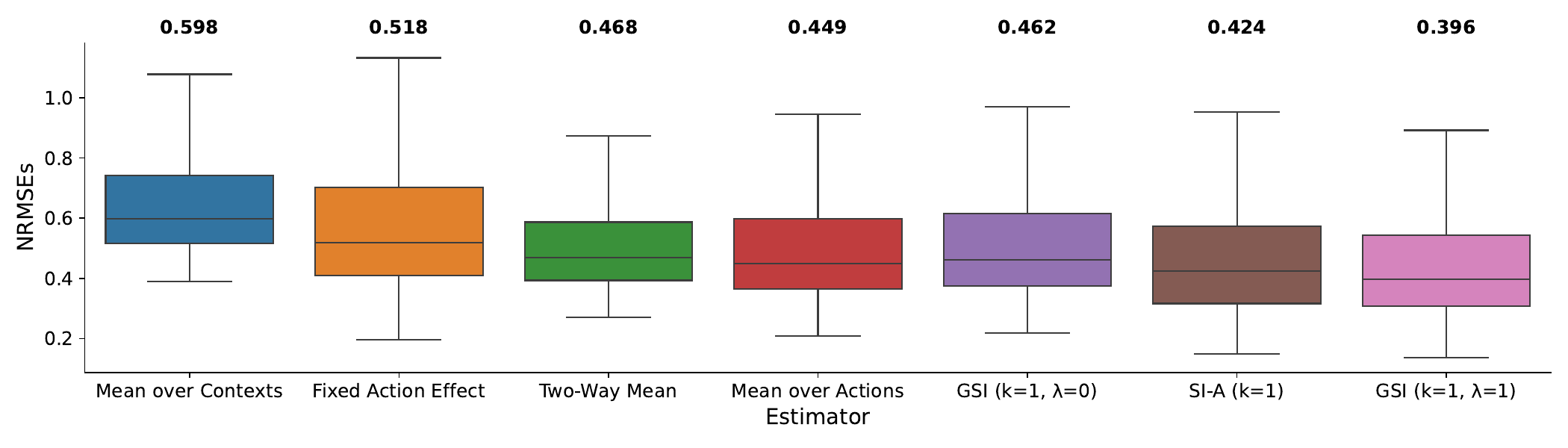}
  \caption{Distribution of NRMSEs for various estimators on the test set ($n=578$). The bold values correspond to the median NRMSE for each estimator.}
  \label{fig:catplot}

\end{figure}

We then test our estimator on data from the CMAP dataset, following the methodology of \cite{squires2022causal}. The CMAP dataset is an incomplete tensor of the interactions of 71 cell types and >20,000 chemical compounds, with each interaction containing the measurement of 978 gene transcription levels. We select two random subsets of 100 compounds (including the "control" compound), the first as validation set and the second as test set.

Without loss of generality, we set $A$ as the set of compounds and $B$ as the set of 71 cell types, yielding two incomplete tensors of size $100 \times 71 \times 978$. Here, as we test the regularized version of $\textsc{GSI}$ whose solution is no longer invariant to linear transformations, we standardize each feature dimension to have mean 0 and standard deviation 1.

For each available interaction, we mask the value and then apply the estimators tested above, comparing their estimate with the ground truth using the normalized root mean-square error (NRMSE) used by \cite{squires2022causal}. We use hyperparameters selected through grid search on the validation set for $\textsc{SI-A}$ (over the values of $k$) and $\textsc{GSI}_\lambda(a, b)$ (over the values of $k, \lambda$). For both, we find that $k=1$ works best (i.e. keep adding compounds until you are left with a single cell type to learn from) and $\lambda$ values between $0.1$ and $2$ performed similarly (with $\lambda=1$ performing slightly better). No regularization ($\lambda=0$) or high regularization ($\lambda > 5$) performed noticeably worse. 

We plot the results on the test set in Figure \ref{fig:catplot}. We find that $\textsc{SI-A}$ and $\textsc{GSI}_1(a, b)$ outperform all baselines with median NRMSEs of $0.424$ and $0.396$ respectively. We then perform a Wilcoxon signed-rank test (\cite{wilcoxon1992individual}) using paired NRMSEs for the two estimators. For the one-sided test that the NRMSEs of $\textsc{GSI}_1(a, b)$ are lower than those of $\textsc{SI-A}$ we obtain a \textbf{p-value of $\mathbf{4.71 \times 10^{-9}}$}, indicating a statistically significant improvement from our estimator.

While $\textsc{GSI}$ does require tuning an additional hyperparameter and is computationally more expensive than $\textsc{SI-A}$, it is computationally tractable and still orders of magnitude faster than other methods (e.g. MissForest (\cite{stekhoven2012missforest}) which, as per (\cite{squires2022causal}) can take hours per prediction in this context).

\section{Conclusion}
In this work, we build upon the work of \cite{squires2022causal} to develop $\textsc{GSI}$, a new causal imputation estimators. We prove the identifiability of this estimator under mild assumptions and show that it outperform baselines on synthetic and CMAP data, thus demonstrating its flexibility and effectiveness. Potential future avenues of research include extending these estimators to the nonlinear case, as well as testing it on larger and more diverse datasets.

\clearpage
\section*{Acknowledgements}
This research was enabled in part by compute resources provided by Mila. MJ and VS are supported by Canada CIFAR AI Chairs. TJ is supported by a FRQNT Master's Training Scholarship.

\bibliography{references}
\bibliographystyle{apalike} 

\clearpage

\end{document}